%% file: 00-main.tex
\documentclass{amia}
\usepackage{url}
\usepackage{graphicx,subfigure,amssymb,amsmath,algorithm,algorithmic,float,multirow}
\usepackage{cases}

\begin{document}
%

\title{Multi-View Graph Convolutional Network and Its Applications on Neuroimage Analysis for Parkinson's Disease}

\author{Xi Sheryl Zhang$^{1,\star}$, Lifang He$^{1,\star}$, Kun Chen$^2$, Yuan Luo$^3$, Jiayu Zhou$^4$, Fei Wang$^{1,\ast}$}

\institutes{\small
$^1$Department of Healthcare Policy and Research, Weill Cornell Medical College, Cornell University, NY\\
$^2$Department of Statistics, University of Connecticut, CT\\
$^3$Department of Preventive Medicine, Feinberg School of Medicine, Northwestern University, IL\\
$^4$Department of Computer Science and Engineering, Michigan State University, MI\\
$^\star$Equal Contribution. $^\ast$Corresponding author, email: \url{few2001@med.cornell.edu}}

\maketitle

{\bf Abstract}. Parkinson's Disease (PD) is one of the most prevalent neurodegenerative diseases that affects tens of millions of Americans. PD is highly progressive and heterogeneous. Quite a few studies have been conducted in recent years on predictive or disease progression modeling of PD using clinical and biomarkers data. Neuroimaging, as another important information source for neurodegenerative disease, has also arisen considerable interests from the PD community. In this paper, we propose a deep learning method based on Graph Convolutional Networks (GCN) for fusing multiple modalities of brain images in relationship prediction which is useful for distinguishing PD cases from controls. On Parkinson's Progression Markers Initiative (PPMI) cohort, our approach achieved $0.9537\pm 0.0587$ AUC, compared with $0.6443\pm 0.0223$ AUC achieved by traditional approaches such as PCA.

\input{01-intro}
\input{02-prelim}
\input{03-method}
\input{04-exp}
\input{05-conclude}
\input{06-ack}

\small
\bibliographystyle{unsrt}
\bibliography{refs}  

\end{document}

%% file: 01-intro.tex
\section{Introduction}
Parkinson's Disease (PD) \cite{dauer2003parkinson} is one of the most prevalent neurodegenerative diseases, which occur when nerve cells in the brain or peripheral nervous system lose function over time and ultimately die. PD affects predominately dopaminergic neurons in substantia nigra, which is a specific area of the brain. PD is a highly progressive disease, with related symptoms progressing slowly over the years. Typical PD symptoms include bradykinesia, rigidity, and rest tremor, which affect speech, hand coordination, gait, and balance. According to the statistics from National Institute of Environmental Healths (NIEHS), at least 500,000 Americans are living with PD\footnote{\url{https://www.niehs.nih.gov/research/supported/health/neurodegenerative/index.cfm}}. The Centers for Disease Control and Prevention (CDC) rated complications from PD as the 14th cause of death in the United States \cite{kochanek2016deaths}.

The cause of PD remains largely unknown. There is no cure for PD and its treatments include mainly medications and surgery. The progression of PD is highly heterogeneous, which means that its clinical manifestations vary from patient to patient. In order to understand the underlying disease mechanism of PD and develop effective therapeutics, many large-scale cross-sectional cohort studies have been conducted. The Parkinson's Progression Markers Initiative (PPMI) \cite{frasier2010parkinson} is one such example including comprehensive evaluations of early stage (idiopathic) PD patients with imaging, biologic sampling, and clinical and behavioral assessments. The patient recruitment in PPMI is taking place at clinical sites in the United States, Europe, Israel, and Australia. This injects enough diversity into the PPMI cohort and makes the downstream analysis/discoveries representative and generalizable.

Quite a few computational studies have been conducted on PPMI data in recent years. For example, Dinov {\em et al}. \cite{dinov2016predictive} built a big data analytics pipeline on the clinical, biomarker and assessment data in PPMI to perform various prediction tasks. Schrag {\em et al}. \cite{schrag2017clinical} predicted the cognitive impairment of the patients in PPMI with clinical variables and biomarkers. Nalls {\em et al}. \cite{nalls2015diagnosis} developed a diagnostic model with clinical and genetic classifications with PPMI cohort. We also developed a sequential deep learning based approach to identify the subtypes of PD on the clinical variables, biomarkers and assessment data in PPMI, and our solution won the PPMI data challenge in 2016 \cite{ppmichallenge}. These studies provided insights to PD researchers in addition to the clinical knowledge.

So far research on PPMI has been mostly utilizing its clinical, biomarker and assessment information. Another important part but under-utilized part of PPMI is its rich neuroimaging information, which includes Magnetic Resonance Imaging (MRI), functional MRI, Diffusion Tensor Imaging (DTI), CT scans, etc. During the last decade, neuroimaging studies including structural, functional and molecular modalities have also provided invaluable insights into the underlying PD mechanism \cite{politis2014neuroimaging}. Many imaging based biomarkers have been demonstrated to be closely related to the progression of PD. For example, Chen {\em et al}. \cite{chen2014imaging} identified significant volumetric loss in the olfactory bulbs and tracts of PD patients versus controls from MRI scans, and the inverse correlation between the global olfactory bulb volume and PD duration. Different observations have been made on the volumetric differences in substantia nigra (SN) on MRI \cite{oikawa2002substantia,peran2010magnetic}. Decreased Fractional Anisotropy (FA) in the SN is commonly observed in PD patients using DTI \cite{cochrane2013diffusion}. With high-resolution DTI, greater FA reductions in caudal (than in middle or rostral) regions of the SN were identified, distinguishing PD from controls with 100\% sensitivity and specificity \cite{vaillancourt2009high}. One can refer to \cite{saeed2017imaging} for a comprehensive review on imaging biomarkers for PD. Many of these neuroradiology studies are strongly hypothesis driven, based on the existing knowledge on PD pathology.

In recent years, with the arrival of the big data era, many computational approaches have been developed for neuroimaging analysis \cite{pereira2009machine,wernick2010machine,ktena2017distance}. Different from conventional hypothesis driven radiology methods, these computational approaches are typically data driven and hypothesis free -- they derive features and evidences directly from neuroimages and utilize them in the derivation of clinical insights on multiple problems such as brain network discovery \cite{bai2017unsupervised,liu2017unified} and imaging genomics \cite{hariri2003imaging,thompson2010imaging}. Most of these algorithms are linear \cite{ryali2010sparse} or multilinear \cite{sajda2004nonnegative}, and they work on a single modality of brain images.

In this paper, we develop a computational framework for analyzing the neuroimages in PPMI data based on Graph Convolutional Networks (GCN) \cite{defferrard2016convolutional}. Our framework learns pairwise relationships with the following steps.

    \underline{\emph{Graph Construction}}. We parcel the structural MRI brain images of each acquisition into a set of Region-of-Interests (ROIs). Each region is treated as a node on a Brain Geometry Graph (BGG), which is undirected and weighted. The weight associated with each pair of nodes is calculated according to the average distance between the geometric coordinates of them in each acquisition. All acquisitions share the same BGG.
    
    \underline{\emph{Feature Construction}}. We use different brain tractography algorithms on the DTI parts of the acquisitions to obtain different Brain Connectivity Graphs (BCGs), which are used as the features for each acquisition. Each acquisition has a BCG for each type of tractography.
    
    \underline{\emph{Relationship Prediction}}. For each acquisition, we learn a feature matrix from each of its BCG through a GCN. Then all the feature matrices are aggregated through element-wise view pooling. Finally, the feature matrices from each acquisition pair are aggregated into a vector, which is fed into a softmax classifier for relationship prediction.

It is worthwhile to highlight the following aspects of the proposed framework.

    \underline{\emph{Pairwise Learning}}. Instead of performing sample-level learning, we learn pairwise relationships, which is more flexible and weaker (sample level labels can always be transformed to pairwise labels but not vice versa). Importantly, such a pairwise learning strategy can increase the training sample size (because each pair of training samples becomes an input), which is very important to learning algorithms that need large-scale training samples (e.g., deep learning).
    
    \underline{\emph{Nonlinear Feature Learning}}. As we mentioned previously, most of the existing machine learning approaches for neuroimaging analysis are based on either linear or multilinear models, which have a limited capacity of exploring the information contained in neuroimages. We leverage GCN, which is a powerful tool that can explore graph characteristics at a spectrum of frequency bands. This brings our framework more potential to achieve good performance.
    
    \underline{\emph{Multi-Graph Fusion}}. Different from conventional approaches that focus on a single graph (image modality), our framework fuses 1) spatial information on the BGG obtained from the MRI part of each acquisition; 2) the features obtained from different BCGs obtained from the DTI part of each acquisition. This effectively leverages the complementary information scattered in different sources.

%% file: 02-prelim.tex
\section{Methodology}
In this section, we first describe the problem setting and then present the details of our proposed approach. To facilitate the description, we denote scalars by lowercase letters (\emph{e.g.}, $x$), vectors by boldfaced lowercase letters (\emph{e.g.}, $\mathbf{x}$), and matrices by boldface uppercase letters (\emph{e.g.}, $\mathbf{X}$). We also use lowercase letters $i, j$ as indices. We write $x_i$ to denote the $i$−th entry of a vector $\mathbf{x}$, and $x_{i,j}$ the entry with row index $i$ and column index $j$ in a matrix $\mathbf{X}$. All vectors are column vectors unless otherwise specified.

\subsection{Problem Setting}


Suppose we have a population of $N$ acquisitions, where each acquisition is subject-specific and associated with $M$ BCGs obtained from different measurements or views. A BCG can be represented as an undirected weighted graph $G=(V, E)$. The vertex set $V=\{v_1, \cdots, v_n \}$ consists of ROIs in the brain and each edge in $E$ is weighted by a connectivity strength, where $n$ is the number of ROIs. We represent edge weights by an $n \times n$ similarity matrix $\mathbf{X}$ with $x_{i,j}$ denoting the connectivity between ROI $i$ and ROI $j$. We assume that the vertices remain the same while the edges vary with views. Thus, for each subject, we have $M$ BCGs: $
    \{G^{(k)} = (V, E^{(k)})\}_{k=1}^M$. A group of similarity matrices $\{\mathbf{X}^{(1)}, \cdots, \mathbf{X}^{(k)}, \cdots, \mathbf{X}^{(M)}\}$ can be derived.

An undirected weighted BGG $\tilde{G} = (V, E)$ is defined based on the geometric information of the region coordinates, which is a $K$-Nearest Neighbor ($K$-NN) graph. The graph has ROIs as vertices $V=\{v_1, \cdots, v_n \}$, where each ROI is associated with coordinates of its center. Edges are weighted by the Gaussian similarity function of Euclidean distances, \emph{i.e.}, $e(v_i, v_j) = \text{exp} (-\frac{\|v_i - v_j \|^2}{2 \sigma^2})$. We identify the set of vertices $N_i$ that are neighbors to the vertex $v_i$ using $K$-NN, and connect $v_i$ and $v_j$ if $v_i \in N_j$ or if $v_j \in N_i$. An adjacency matrix $\tilde{\mathbf{A}}$ can then be associated with $\tilde{G}$ representing the similarity to nearest similar ROIs for each ROI, with the elements:
\begin{align*}
\tilde{a}_{i, j} =
        \begin{cases}
            e(v_i, v_j), & \text{if}~~v_i \in N_j ~\text{or}~v_j \in N_i  \\
            0, & \text{otherwise}. \nonumber
        \end{cases}
\end{align*}

Our goal is to learn a feature representation for each subject by fusing its BCGs and the shared BGG, which captures both the local traits of each individual subject and the global traits of the population of subjects. Specifically, we develop a customized Multi-View Graph Convolutional Network (MVGCN) model to learn feature representations on neuroimaging data.




%% file: 03-method.tex
\subsection{Our Approach}

\begin{figure}[htb]
    \centering
    \includegraphics[width=4in]{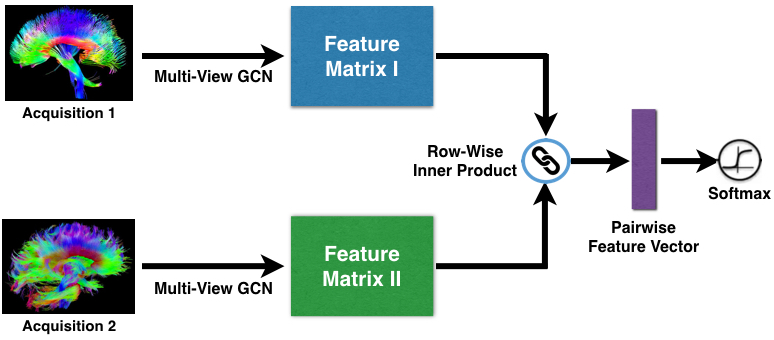}
    \caption{Overall flowchart of our framework.}
    \label{fig:overall}
\end{figure}

\noindent{\bf Overview.} Fig.~\ref{fig:overall} provides an overview of the MVGCN framework we develop for relationship prediction on multi-view brain graphs. Our model is a deep neural network consisting of three main components: the first component is a multi-view GCN for extracting the feature matrices from each acquisition, the second component is a pairwise matching strategy for aggregating the feature matrices from each pair of acquisitions into feature vectors, and the third component is a softmax predictor for relationship prediction. All of these components are trained using back-propagation and stochastic optimization. Note that MVGCN is an end-to-end architecture without extra parameters involved for view pooling and pairwise matching, Also, all branches of the used views share the same parameters in the multi-view GCN component. We next give details of each component.

\noindent{\bf $\bullet$ \emph{C1: Multi-View GCN.}}
Traditional convolutional neural networks (CNN) rely on the regular grid-like structure with a well-defined neighborhood at each position in the grid (\emph{e.g.} 2D and 3D images). On a graph structure there is usually no natural choice for an ordering of the neighbors of a vertex, therefore it is not trivial to generalize the convolution operation to the graph setting. Shuman et al. showed that this generalization can be made feasible by defining graph convolution in the spectral domain and proposed a GCN. Motivated by the fact that GCN can effectively model the nonlinearity of samples in a population and has superior capability to explore graph characteristics at a spectrum of frequency bands, we propose a multi-view GCN for an effective fusion of populations of graphs with different views. It consists of two fundamental steps: (i) the design of convolution operator on multiple graphs across views, (ii) a view pooling operation that groups together multi-view graphs.


\noindent{\bf \emph{Graph Convolution.}} An essential point in GCN is to define graph convolution in the spectral domain based on Laplacian matrix and graph Fourier transform (GFT). We consider the normalized graph Laplacian $
    \mathbf{L}=\mathbf{I}-\mathbf{D}^{-1/2}\mathbf{A}\mathbf{D}^{-1/2}$,
where $\mathbf{A} \in \mathbb{R}^{n \times n} $ is the adjacency matrix associated with the graph, $\mathbf{D} \in \mathbb{R}^{n \times n}$ is the diagonal degree matrix with $d_{i,i} = \sum_j a_{i,j} $, and $\mathbf{I} \in \mathbb{R}^{n \times n}$ is the identity matrix. As $\mathbf{L}$ is a real symmetric positive semidefinite matrix, it can be decomposed as $\mathbf{L} = \mathbf{U}\boldsymbol{\Lambda}\mathbf{U}^{\mathrm{T}}$, where $\mathbf{U} \in \mathbb{R}^{n \times n}$ is the matrix of eigenvectors with $\mathbf{U}\mathbf{U}^\mathrm{T} = \mathbf{I}$ (referred to as the Fourier basis) and $\boldsymbol{\Lambda} \in \mathbb{R}^{n \times n}$ is the diagonal matrix of eigenvalues $\{\lambda_i \}_{i=1}^n$. The eigenvalues represent the frequencies of their associated eigenvectors, $\emph{i.e.}$ eigenvectors associated with larger eigenvalues oscillate more rapidly between connected vertices. Specifically, in order to obtain a unique frequency representation for the signals on the set of graphs, we define the Laplacian matrix on the BGG $\tilde{G}$, as all graphs share a common structure with adjacency matrix $\tilde{\mathbf{A}}$.

Let $\mathbf{x} \in \mathbb{R}^n$ be a signal defined on the vertices of a graph $G$, where $x_i$ denotes the value of the signal at the $i$-th vertex. The GFT is defined as $\hat{\mathbf{x}} = \mathbf{U}^{\mathrm{T}}\mathbf{x}$, which converts signal $\mathbf{x}$ to the spectral domain spanned by the Fourier basis $\mathbf{U}$. Then the graph convolution can be defined as:
\begin{equation} \label{eq:GC}
	\mathbf{y} = g_\theta(\mathbf{L}) \mathbf{x}  = g_\theta( \mathbf{U\Lambda U}^{\mathrm{T}} ) \mathbf{x} = \mathbf{U} g_\theta(\mathbf{\Lambda}) \mathbf{U}^{\mathrm{T}}\mathbf{x},
\end{equation}
where $\theta$ is a vector of Fourier coefficients to be learned, and $g_\theta$ is called the filter which can be regarded as a function of $\mathbf{\Lambda}$. To render the filters $s$-localized in space and reduce the computational complexity, $g_\theta$ can be approximated by a truncated expansion in terms of Chebyshev polynomials of order $s$ \cite{defferrard2016convolutional}. That is,
\begin{equation} \label{eq:Chebyshev}
	g_\theta(\mathbf{\Lambda})  = \sum\nolimits_{p = 0}^{s-1} \theta_p T_p (\tilde{\mathbf{\Lambda}}),
\end{equation}
where the parameter $\theta_{p} \in \mathbb{R}^s$ is a vector of Chebyshev coefficients and $T_p (\tilde{\mathbf{\Lambda}}) \in \mathbb{R}^{n \times n} $ is the Chebyshev polynomial of order $s$ evaluated at $\tilde{\mathbf{\Lambda}} = 2 \mathbf{\Lambda}/\lambda_{\max} - \mathbf{I}$, a diagonal matrix of scaled eigenvalues that lies in $[-1, 1]$.

Substituting Eq.~(\ref{eq:Chebyshev}) into Eq.~(\ref{eq:GC}) yields
$\mathbf{y} = g_\theta (\mathbf{L}) \mathbf{x} = \sum\nolimits_{p = 0}^{s-1} \theta_p T_p(\tilde{\mathbf{L}}) \mathbf{x}$,
where $\tilde{\mathbf{L}} = \frac{2}{\lambda_{\max}} \mathbf{L} - \mathbf{I}$.
Denoting $\tilde{\mathbf{x}}_p = T_p(\tilde{\mathbf{L}}) \mathbf{x}$, we can use the recurrence relation to compute $\tilde{\mathbf{x}}_i = 2\tilde{\mathbf{L}} \tilde{\mathbf{x}}_{p-1} - \tilde{\mathbf{x}}_{p-2} $ with $\tilde{\mathbf{x}}_0 = \mathbf{x}$ and $\tilde{\mathbf{x}}_1 = \tilde{\mathbf{L}} \mathbf{x}$. Finally, the $j$th output feature map in a GCN is given by:
\begin{equation}
\mathbf{y}_j = \sum\nolimits_{i=1}^{F_{in}} g_{\theta_{i,j}} (\mathbf{L}) \mathbf{x}_{i}, 
\end{equation}
yielding $F_{in} \times F_{out}$ vectors of trainable Chebyshev coefficients $\theta_{i,j} \in \mathbb{R}^s$, where $\mathbf{x}_i$ denotes the input feature maps from a graph. For each BCG, $\mathbf{x}_i$ corresponds to the $i$-th row of the respective input connectivity matrix $\mathbf{X}$, and the initial $F_{in}$ is $n$ which equals to the number of brain ROIs. The outputs are collected into a feature matrix $\mathbf{Y} = [\mathbf{y}_1, \mathbf{y}_2, \cdots, \mathbf{y}_{F_{out}}] \in \mathbb{R}^{n \times F_{out}}$, where each row represents the extracted features of an ROI.

\begin{figure}[tb]
    \centering
    \includegraphics[width=4in]{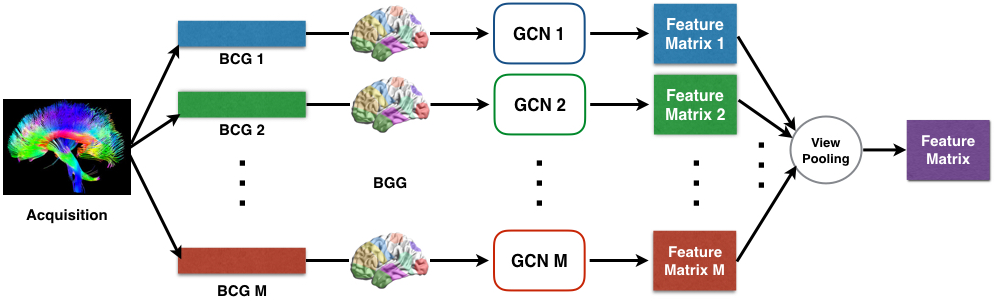}
    \caption{The flowchart of the multi-view GCN component.}
    \label{fig:gcn}
\end{figure}

\noindent{\bf \emph{View Pooling.}} For each subject, the output of GCN are $M$ feature matrices $\{\mathbf{Y}^{(1)}, \cdots, \mathbf{Y}^{(k)}, \cdots, \mathbf{Y}^{(M)}\}$, where each matrix $\mathbf{Y}^{(k)} \in \mathbb{R}^{n \times F_{out}} $ corresponds to a view. Similar to the view pooling layer in the multi-view CNN \cite{su2015multi}, we use element-wise maximum operation across all $M$ feature matrices in each subject to aggregate multiple views together, producing a shared feature matrix $\mathbf{Z}$. An alternative is an element-wise mean operation, but it is not as effective in our experiments (see Table \ref{table:multi view comparisons}). The reason might be that the maximum operation learns to combine the views instead of averaging, and thus can use the more informative views of each feature while ignoring others. 

Fig.~\ref{fig:gcn} gives the flowchart of our multi-view GCN. Based on this multi-view GCN, different views of BCGs can be progressively fused in accordance with their similarity matrices, which can capture both local and global structural information from BCGs and BGG.

\noindent{\bf $\bullet$ \emph{C2: Pairwise Matching.}} Training deep learning model requires a large amount of training data, but usually very few data are available from clinical practice. We take advantage of the pairwise relationships between subjects to guide the process of deep learning \cite{koch2015siamese,ktena2017distance}. Similarity is an important type of pairwise relationship that measures the relatedness of two subjects. The basic assumption is that, if two subjects are similar, they should have a high probability to have the same class label. 

Let $\mathbf{Z}_p$ and $\mathbf{Z}_q$ be the feature matrices for any subject pair obtained from multi-view GCN, we can use them to compute an ROI-ROI similarity score. To do so, we first normalize each matrix so that the sum of squares of each row is equal to 1, and then define the following pairwise similarity measure using the row-wise inner product operator:
\begin{equation} \label{eq:pairwise}
    \text{sim}(\mathbf{z}_p^i, \mathbf{z}_q^i) = {\mathbf{z}^i_p}^\mathrm{T} \mathbf{z}_q^i,~~ i=1, 2, \cdots, n.
\end{equation}
where $\mathbf{z}_p^i$ and $\mathbf{z}_q^i$ are the $i$-th row vectors of the normalized matrices $\mathbf{Z}_p$ and $\mathbf{Z}_q$, respectively.

\noindent{\bf $\bullet$ \emph{C3: Softmax}}. For each pair, the output of the pairwise matching layer is a feature vector $\mathbf{r}$, where each element is given by Eq.~(\ref{eq:pairwise}). Then, this representation is passed to a fully connected softmax layer for classification. It computes the probability distribution over the labels:
\begin{align}\label{eq:softmax}
    p(y=j|\mathbf{r}) = {\text{exp} (\mathbf{w}^{\mathrm{T}}_j \mathbf{r})}/{\left[\sum\nolimits_{c=1}^C \text{exp} (\mathbf{w}^{\mathrm{T}}_c \mathbf{r})\right]},
\end{align}
where $\mathbf{w}_c$ is the weight vector of the $c$-th class, and $\mathbf{r}$ is the final abstract representation of the input example obtained by a series of transformations from the input layer through a series of convolution and pooling operations.




%% file: 04-exp.tex
\section{Experiments and Results}
In order to evaluate the effectiveness of our proposed approach, we conduct extensive experiments on real-life Parkinson’s Progression Markers Initiative (PPMI) data for relationship prediction and compare with several state-of-the-art methods. In the following, we introduce the datasets used and describe details of the experiments. Then we present the results as well as the analysis.

\textbf{Data Description.} We consider the DTI acquisition on $754$ subjects, where $596$ subjects are Parkinson's Disease (PD) patients and the rest $158$ are Healthy Control (HC) ones. Each subject's raw data were aligned to the b0 image using the FSL\footnote{\url{http://www.fmrib.ox.ac.uk/fsl}} eddy-correct tool to correct for head motion and eddy current distortions. The gradient table is also corrected accordingly. Non-brain tissue is removed from the diffusion MRI using the Brain Extraction Tool (BET) from FSL. To correct for echo-planar induced (EPI) susceptibility artifacts, which can cause distortions at tissue-fluid interfaces, skull-stripped b0 images are linearly aligned and then elastically registered to their respective preprocessed structural MRI using Advanced Normalization Tools (ANTs\footnote{\url{http://stnava.github.io/ANTs/}}) with SyN nonlinear registration algorithm. The resulting 3D deformation fields are then applied to the remaining diffusion-weighted volumes to generate full preprocessed diffusion MRI dataset for the brain network reconstruction. In the meantime, 84 ROIs are parcellated from T1-weighted structural MRI using Freesufer\footnote{\url{https://surfer.nmr.mgh.harvard.edu}} and each ROI's coordinate is defined using the mean coordinate for all voxels in that ROI.

Based on these 84 ROIs, we reconstruct six types of BCGs for each subject using six whole brain tractography algorithms, including four tensor-based deterministic approaches: Fiber Assignment by Continuous Tracking (FACT) \cite{mori1999three}, the 2nd-order Runge-Kutta (RK2) \cite{basser2000vivo}, interpolated streamline (SL) \cite{conturo1999tracking}, the tensorline (TL) \cite{lazar2003white}, one Orientation Distribution Function (ODF)-based deterministic approach \cite{aganj2010reconstruction}: ODF-RK2 and one ODF-based probabilistic approach: Hough voting \cite{aganj2011hough}. Please refer to \cite{zhan2015comparison} for the details of whole brain tractography computations. Each resulted network for each subject is $84 \times 84$. To avoid computation bias in the later feature extraction and evaluation sections, we normalize each brain network by the maximum value in the matrix, as matrices derived from different tractography methods have different scales and ranges.

\textbf{Experimental Settings.} To learn similarities between graphs, brain networks in the same group (PD or HC) are labeled as matching pairs while brain networks from different groups are labeled as non-matching pairs. Hence, we have $283,881$ pairs in total, with $189,713$ matching samples and $94,168$ non-matching samples. $5$-fold cross validation is adopted in all of our experiments by separating brain graphs into $5$ stratified randomized sets, where $1$-fold held-out data used for the testing sample pairs and the rest used for training. Using the coordinate information of ROIs in DTI, we construct a $10$-NN BGG in our method, which has $84$ vertices and $527$ edges. For graph convolutional layers, the order of Chebyshev polynomials $s=30$ and the output feature dimension $F_{out}=128$ are used. For fully connected layers, the number of feature dimensions is $1024$ in the baseline of one fully connected layer, and those are set as $1024$ and $64$ for the baseline of two layers. The Adam optimizer~\cite{kingma2014adam} is used with the initial learning rate $0.005$. The above parameters are optimal settings for all the methods by performing cross-validation. MVGCN code and scripts are available on a public repository (\url{https://github.com/sheryl-ai/MVGCN}).            

\begin{table}[]
\small
\centering
\renewcommand\arraystretch{1.2}
\caption{Results for classifying matching vs. non-matching brain networks in terms of AUC metric.}
\label{table:single view comparisons}
\begin{tabular}{lcccccc}
\hline
\multicolumn{1}{c}{\multirow{2}{*}{\textbf{Methods}}} & \multicolumn{6}{c}{\textbf{Modals}} \\ \cline{2-7} 
\multicolumn{1}{c}{} &\textbf{FACT} &\textbf{RK2} &\textbf{SL} &\textbf{TL} &\textbf{ODF-RK2} &\textbf{Hough}      \\ 
\hline
Raw Edges     &58.47 (4.05) &62.54 (6.88) &59.39 (5.99)   &61.94 (5.00)  &60.93 (5.60) &64.49 (3.56)             \\
PCA         &64.10 (2.10)   &63.40 (2.72) &64.43 (2.23)            &62.46 (1.46)   &60.93 (2.63) &63.46 (3.52)           \\
FCN  &66.17 (2.00) &65.11 (2.63) &65.00 (2.29) &64.33 (3.34) &68.80 (2.80) &61.91 (3.42)      \\
FCN$_{2l}$   &82.36 (1.87) &81.02 (4.28) &81.68 (2.49) &81.99 (3.44) &82.53 (4.74) &81.77 (3.74)   \\ 
GCN    &\textbf{92.67 (4.94)} &\textbf{92.99 (4.95)} &\textbf{92.68 (5.32)} &\textbf{93.75 (5.39)} &\textbf{93.04 (5.26)} &\textbf{93.90 (5.48)}       \\
\hline
\end{tabular}
\end{table}

\begin{table}[]
\small
\centering
\tabcolsep 0.3in
\renewcommand\arraystretch{1.2}
\caption{Comparison of binary classification (AUC) and acquisition clustering (NMI) results using both single-view and multi-view architectures.}
\label{table:multi view comparisons}
\begin{tabular}{lcc}
\hline
Architectures            & AUC             & NMI    \\ 
\hline
PCA100-M-S              &64.43 (2.23)     &0.39\\
FCN1024-M-FCN64-S          &82.53 (4.74)     &0.87      \\ 
GCN128-M-S       &93.75 (5.39)     &0.98 \\ 
\hline
MVGCN128-M-S$_{mean}$ &94.74 (5.62)  &1.00      \\ 
MVGCN128-M-S$_{max}$  & \textbf{95.37 (5.87)} &1.00      \\ 
\hline
\end{tabular}
\end{table}

\begin{figure}[!t]
\centering
\subfigure[PCA]{
\begin{minipage}[b]{0.32\linewidth}
\centering
\includegraphics[width=\textwidth]{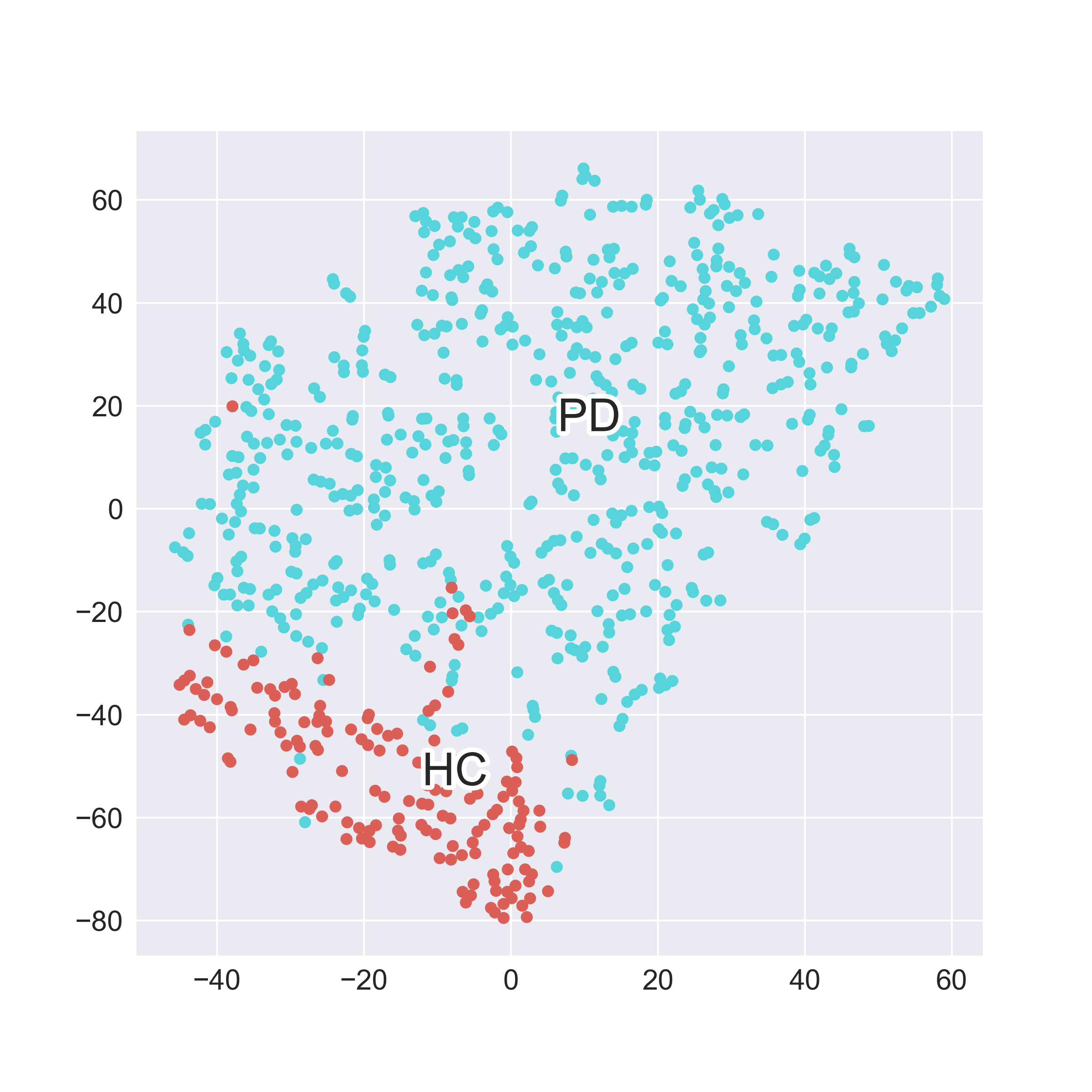}
\end{minipage}}%
\subfigure[FCN$_{2l}$]{
\begin{minipage}[b]{0.32\linewidth}
\centering
\includegraphics[width=\textwidth]{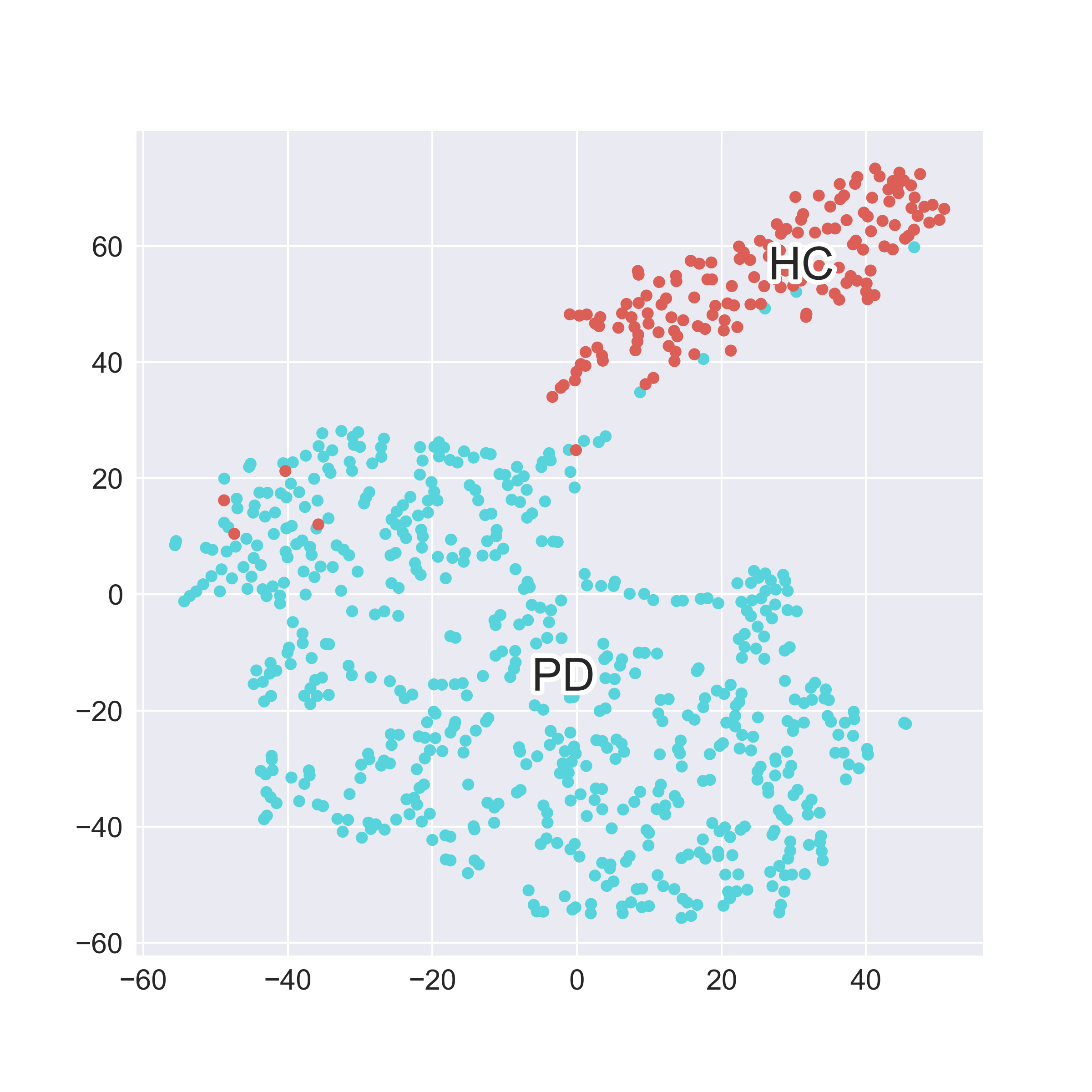}
\end{minipage}}%
\subfigure[MVGCN]{
\begin{minipage}[b]{0.32\linewidth}
\centering
\includegraphics[width=\textwidth]{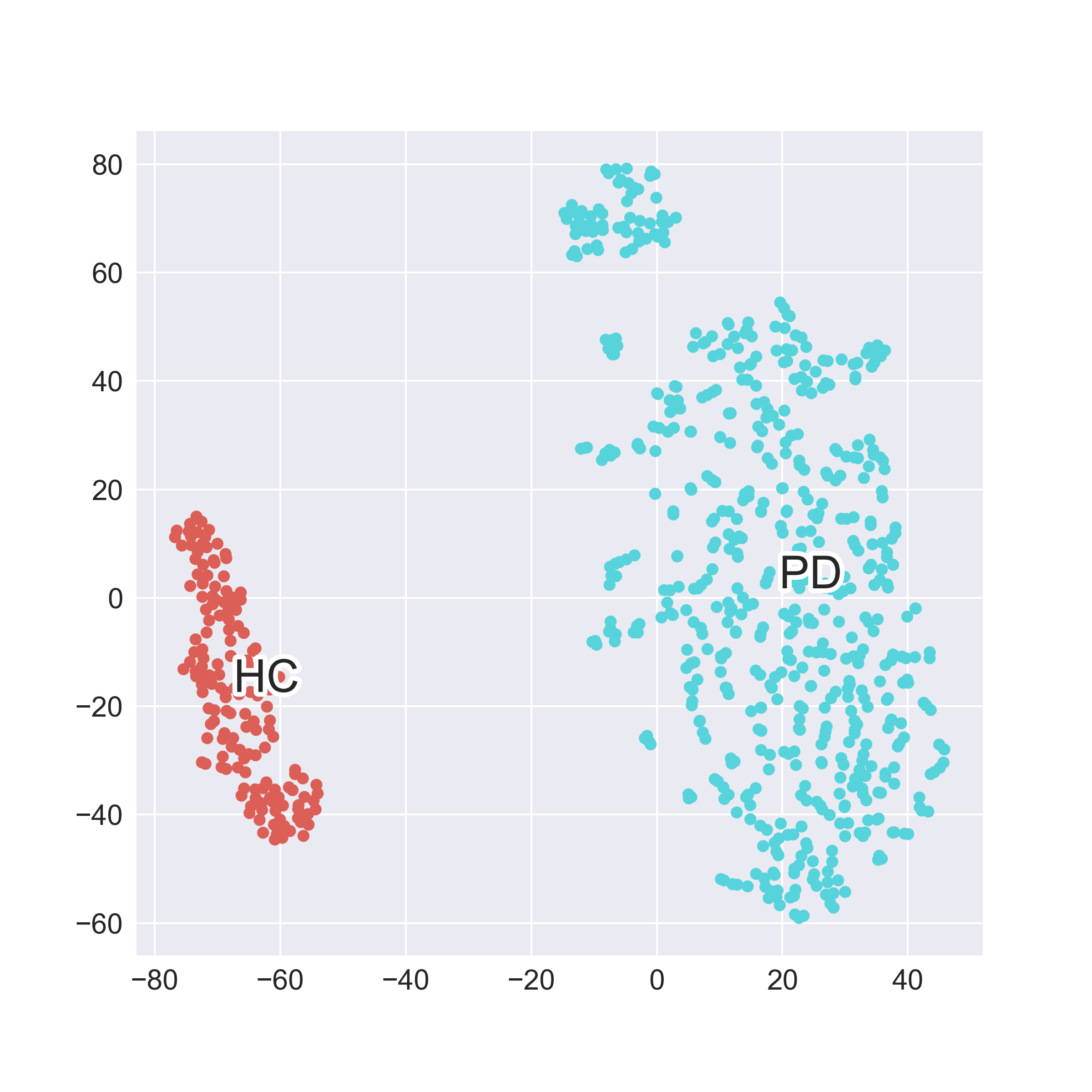}
\end{minipage}}
\vspace{-0.5cm}
\caption{Visualization of the DTI acquisition clusters. The acquisitions are mapped to the 2D space using the t-SNE algorithm with the predicted values of pairwise relationship as input. Blue denotes PD, Red denotes HC.}
\label{figure:visualization}
\end{figure}

\begin{figure}[t]
\centering
\subfigure[Top-10 similar ROI for PD group]{
\begin{minipage}[b]{0.45\linewidth}
\centering
\includegraphics[width=\textwidth]{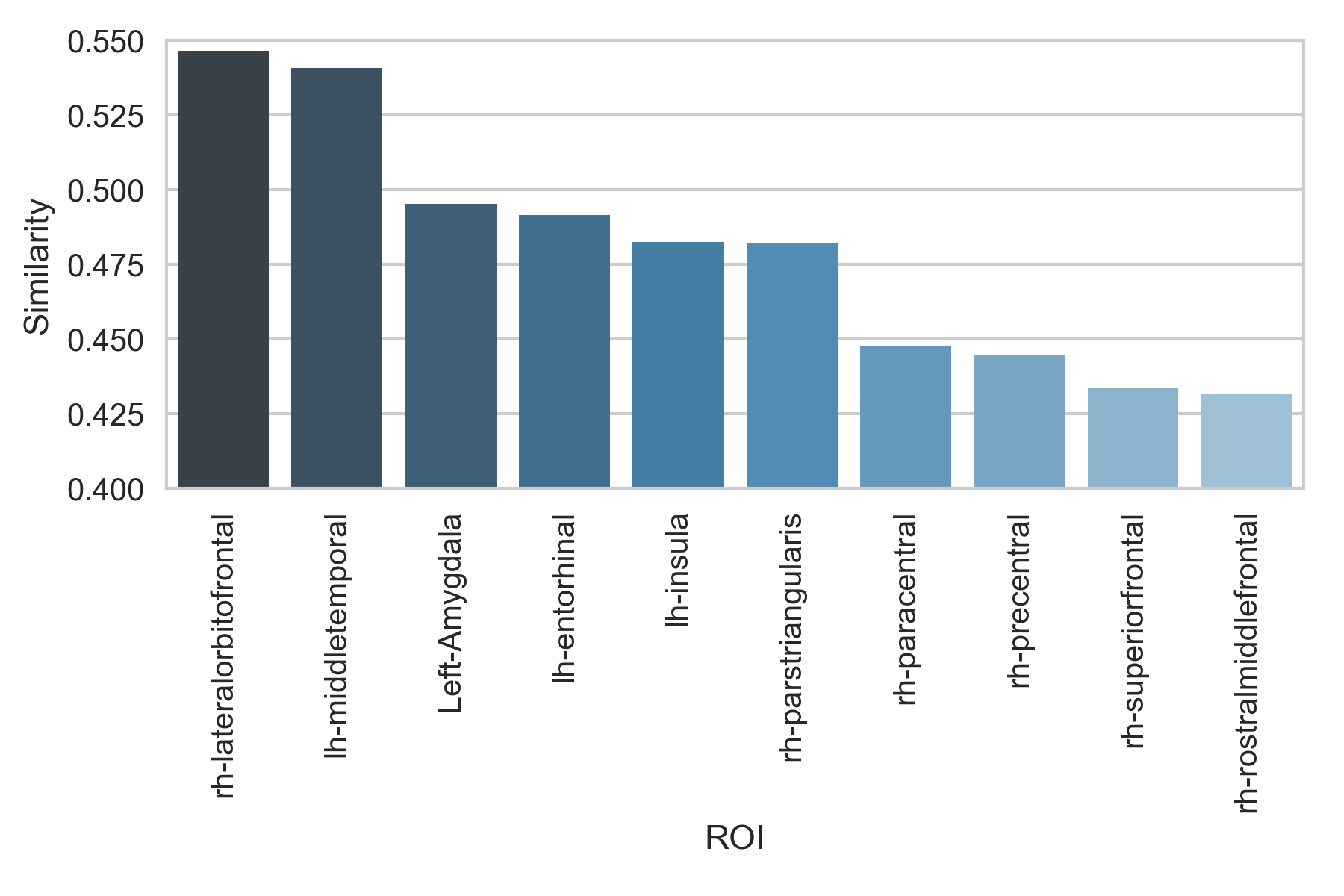}
\end{minipage}}%
\subfigure[Top-10 similar ROI for HC group]{
\begin{minipage}[b]{0.45\linewidth}
\centering
\includegraphics[width=\textwidth]{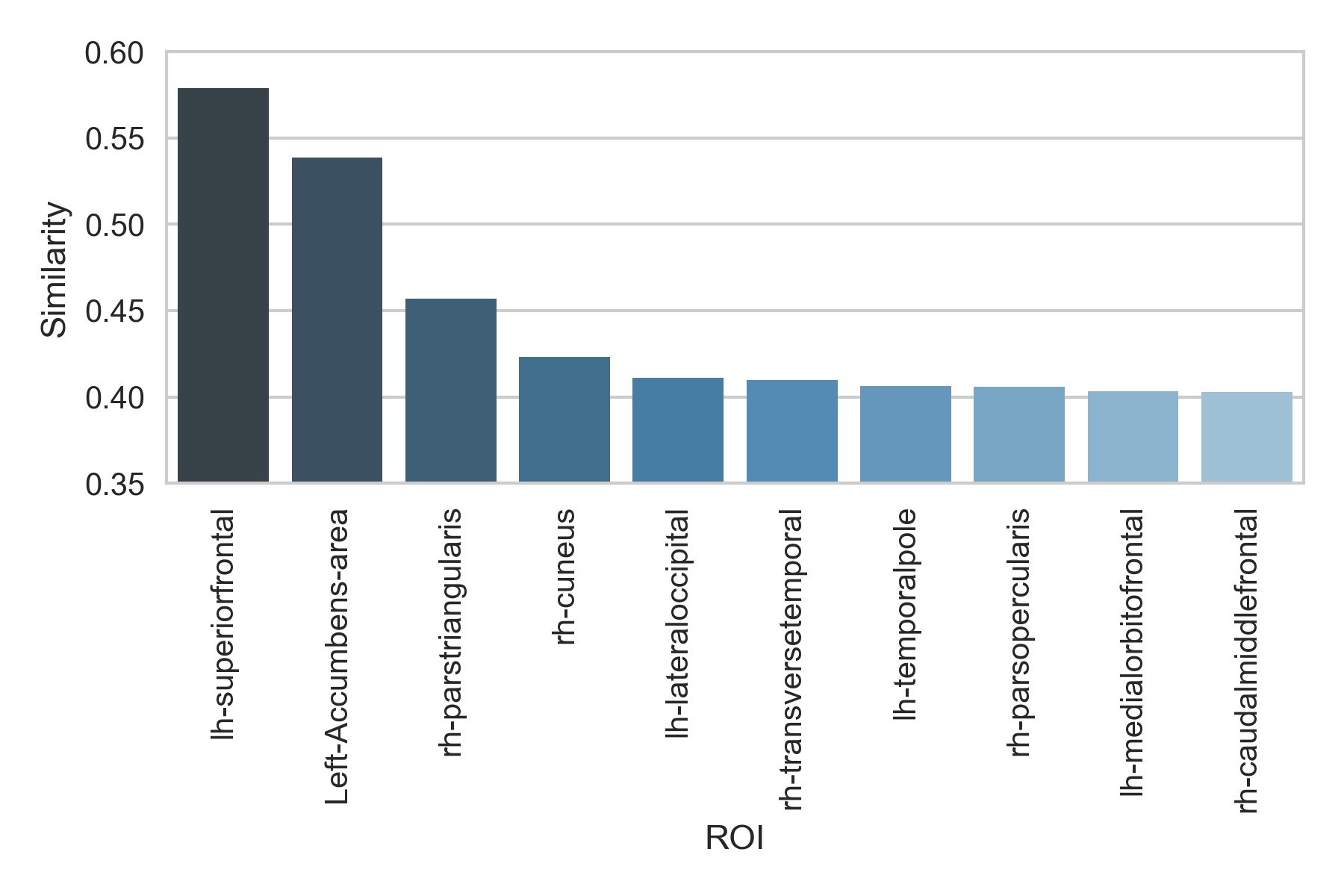}
\end{minipage}}
\subfigure[Top-10 similar ROI between PD and HC]{
\begin{minipage}[b]{0.45\linewidth}
\centering
\includegraphics[width=\textwidth]{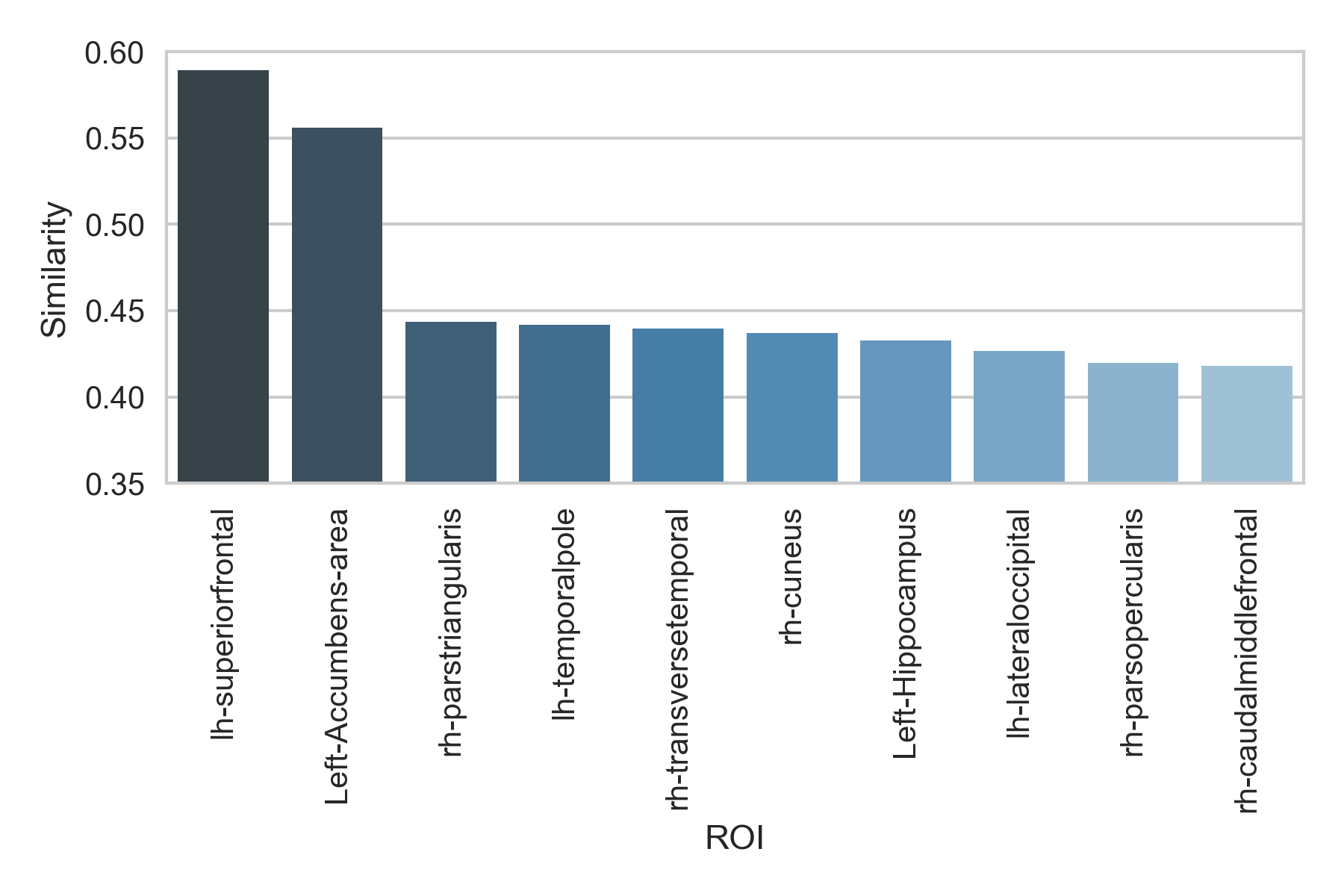}
\end{minipage}}%
\subfigure[Top-10 dissimilar ROI between PD and HC]{
\begin{minipage}[b]{0.45\linewidth}
\centering
\includegraphics[width=\textwidth]{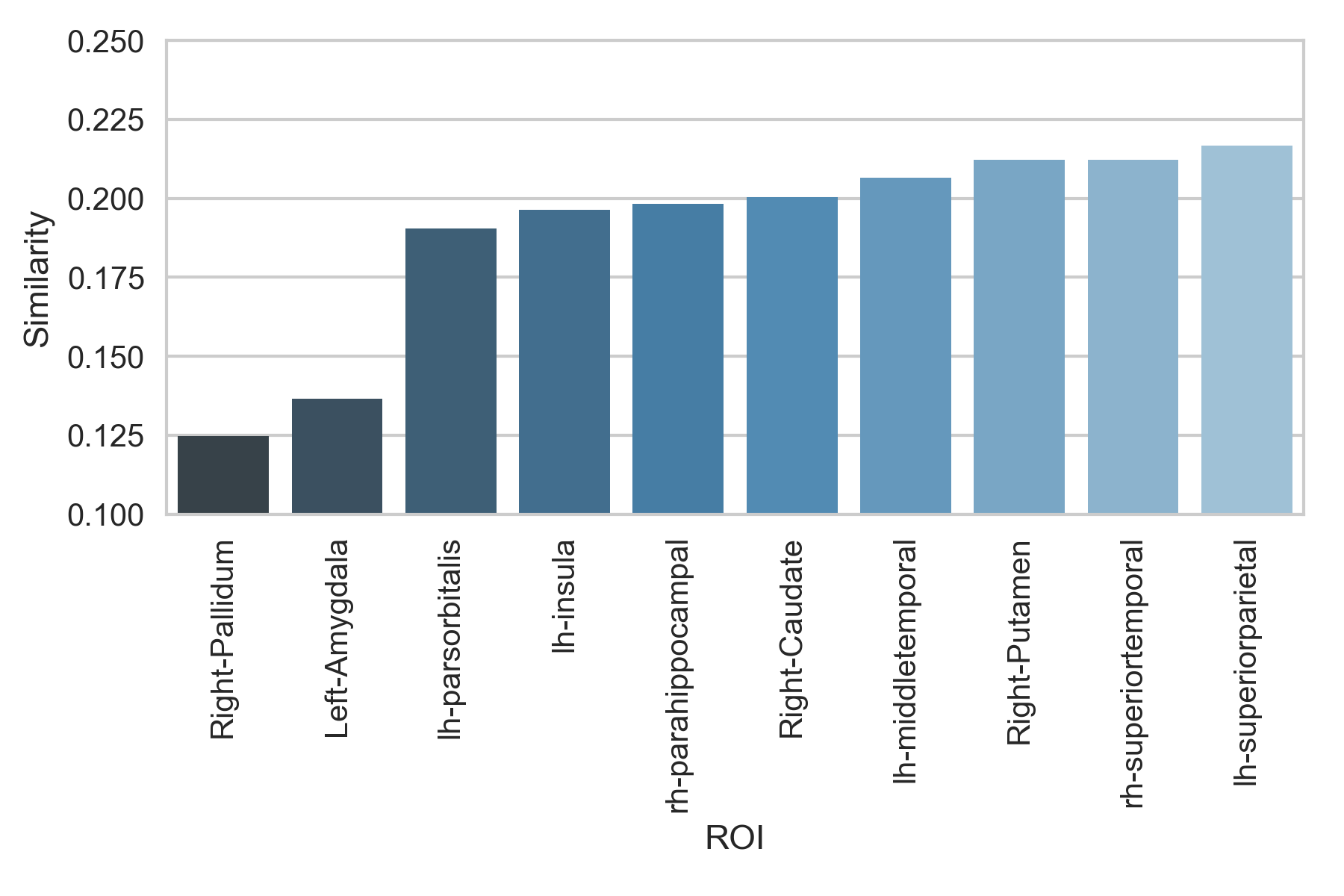}
\end{minipage}}
\vspace{-0.5cm}
\caption{Visualization of the learned ROI-ROI similarities from averaged pairwise feature vectors in the certain groups. Top-$10$ similar or dissimilar ROIs for PD and HC groups are shown in (a)-(d) respectively.}
\label{figure:roi-visualization}
\end{figure}

\textbf{Results.} Since our target is to predict relations (matching vs. non-matching) between pairwise BCGs, the performance of binary classification are evaluated using the metric of Area Under the Curve (AUC). Table~\ref{table:single view comparisons} provides the results of individual views using the following methods: raw edges-weights, PCA, feed-forward fully connected networks (FCN and FCN$_{2l}$), and graph convolutional network (GCN), where FCN$_{2l}$ is a two-layer FCN. Through the compared methods, the feature representation of each subject in pairs can be learned. For a fair comparison, pairwise matching component and software component are utilized for all the methods. The best performance of GCN-based method achieves an AUC of $93.90\%$. It is clear that GCN outperforms the raw edges-weights, conventional linear dimension reduction method PCA and nonlinear neural networks FCN and FCN$_{2l}$.    

Table~\ref{table:multi view comparisons} reports the performance on classification and acquisition clustering of our proposed MVGCN with three baselines. The architectures of neural networks by the output dimensions of the corresponding hidden layers are presented. M denotes the matching layer based on Eq.~(\ref{eq:pairwise}), S denotes the softmax operation in Eq.~(\ref{eq:softmax}). The numbers denote the dimensions of extracted features at different layers. For our study, we evaluate both element-wise max pooling and mean pooling in the view pooling component. Specifically, to test the effectiveness of the learned similarities, we also evaluate the clustering performance in terms of Normalized Mutual Information (NMI). The acquisition clustering algorithm we used is $K$-means ($K = 2$, PD and HC). The results show that our MVGCN outperforms all baselines on both classification and acquisition clustering tasks, with an AUC of $95.37\%$ and an NMI of $1.00$.

In order to test whether the prediction results are meaningful for distinguishing brain networks as PD or HC, we visualize the Euclidean distance for the given 754 DTI acquisitions. Since the output values of all the matching models can indicate the pairwise similarities between acquisitions, we map it into a 2D space with t-SNE \cite{maaten2008visualizing}. Fig.~\ref{figure:visualization} compares the visualization results with different approaches. The feature extraction by PCA cannot separate the PD and HC perfectly. The result of FCN$_{2l}$ in the view ODF-RK2 that has the best AUC is much better, and two clusters can be observed with a few overlapped acquisitions. Compared with PCA and FCN$_{2l}$, the visualization result of MVGCN with max view pooling clearly shows two well-separated and relatively compact groups.

Furthermore, we investigate the extracted pairwise feature vectors of the proposed MVGCN. After the ROI-ROI based pairwise matching, the output for each pair is a feature vector embedding the similarity of the given two acquisitions, with each element associated with a ROI. By visualizing the value distribution over ROIs, we can interpret the learned pairwise feature vector of our model. Fig.~\ref{figure:roi-visualization} reports the most $10$ similar or dissimilar ROI for PD or HC groups. The similarities are directly extracted from the output representations of the pairwise matching layer. We compute the averaged values of certain groups. For instance, the similarity distributions are computed given the pairwise PD samples, and the values of the top-$10$ ROI are shown in Fig.~\ref{figure:roi-visualization}(a). According to the results, lateral orbitofrontal area, middle temporal and amygdala areas are the three most similar ROIs for PD patients, while important ROIs such as caudate and putamen areas are discriminative to distinguish PD and HC (see Fig.~\ref{figure:roi-visualization}(d)). The observations demonstrate that the learned pairwise feature vectors are consistent with some clinical discoveries~\cite{gao2016csf} and thus verify the effectiveness of the MVGCN for neuroimage analysis. 

%% file: 05-conclude.tex
\section{Discussion}
The underlying rationale of the proposed method is modeling the multiple brain connectivity networks (BCGs) and a brain geometry graph (BGG) based on the common ROI coordinate simultaneously. Since BCGs are non-Euclidean, it is not straightforward to use a standard convolution that has impressive performances on the grid. Additionally, multi-view graph fusion methods~\cite{ma2018drug} allow us to explore various aspects of the given data. Our non-parametric view pooling is promising in practice. Furthermore, the pairwise learning strategies can satisfy the ``data hungry'' neural networks with few acquisitions~\cite{koch2015siamese}. Our work has demonstrated strong potentials of graph neural networks on the scenario of multiple graph-structured neuroimages. Meanwhile, the representations learned by our approach can be straightforwardly interpreted. 
However, there are still some limitations. The current approach is completely data-driven without utilization of any clinical domain knowledge. The clinical data such as Electronic Health Records are not considered in the analysis of the disease. In the future, we will continue our research specifically along these directions.

\section{Conclusion}
\label{sec:conclusion}
We propose a multi-view graph convolutional network method called MVGCN in this paper, which can directly take brain graphs from multiple views as inputs and do prediction on that. We validate the effectiveness of MVGCN on real-world Parkinson's Progression Markers Initiative (PPMI) data for predicting the pairwise matching relations. We demonstrate that our proposed MVGCN can not only achieve good performance, but also discover interesting predictive patterns.

%% file: 06-ack.tex
\section*{Acknowledgement}
The work is supported by NSF IIS-1716432 (FW), NSF IIS-1650723 (FW), NSF IIS-1750326 (FW), NSF IIS-1718798 (KC), and MJFF14858 (FW). Data used in the preparation of this article were obtained from the Parkinson's Progression Markers Initiative (PPMI) database (\url{http://www.ppmi-info.org/data}). For up-to-date information on the study, visit \url{http://www.ppmi-info.org}. PPMI -- a public-private partnership -- is funded by the Michael J. Fox Foundation for Parkinson's Research and funding partners, including Abbvie, Avid, Biogen, Bristol-Mayers Squibb, Covance, GE, Genentech, GlaxoSmithKline, Lilly, Lundbeck, Merk, Meso Scale Discovery, Pfizer, Piramal, Roche, Sanofi, Servier, TEVA, UCB and Golub Capital. The authors would like to thank the support from Amazon Web Service Machine Learning for Research Award (AWS MLRA).